\pdfoutput=1

\documentclass[11pt]{article}
\usepackage{naacl2024}

\usepackage{times}
\usepackage{latexsym}
\usepackage{graphicx}
\usepackage{amsmath}
\usepackage{amssymb}
\usepackage{mathabx}
\usepackage{xcolor}
\usepackage{tabularx}
\usepackage{array}
\usepackage{caption}
\usepackage{subcaption}
\usepackage{soul}
\usepackage{url}
\usepackage{hyperref}
\usepackage{colortbl}
\usepackage{pifont}

\newcommand{\grayline}{\arrayrulecolor{lightgray}\hline\arrayrulecolor{black}}

\definecolor{blue}{HTML}{3E74D1}
\definecolor{red}{HTML}{E22146}
\definecolor{green}{HTML}{70b642}
\definecolor{violet}{HTML}{af649b}
\definecolor{lightgray}{HTML}{c6c6c6}
\newcommand{\blue}[1]{\textcolor{blue}{#1}}
\newcommand{\red}[1]{\textcolor{red}{#1}}

\newcommand{\violet}[1]{\textcolor{violet}{#1}}

\usepackage[T1]{fontenc}

\usepackage[utf8]{inputenc}

\usepackage{linguex}
\usepackage{microtype}
\usepackage{inconsolata}
\usepackage{multicol}
\usepackage{booktabs}

\definecolor{customblue}{RGB}{180, 225, 249}
\definecolor{bluetext}{RGB}{0, 118, 186}
\sethlcolor{customblue}

\newcommand{\rulesep}{\color{lightgray}{\unskip\ \vrule \ }}

\title{In-context Learning Generalizes, But Not Always Robustly:\\The Case of Syntax}

\author{Aaron Mueller$^{1,2}$\thanks{\ \ \ Parts of this work completed when A.M. was a long-term visitor at New York University.}\ \ \ \ \ Albert Webson$^3$\ \ \ \ \ Jackson Petty$^4$\thanks{\ \ \ Work completed while J.P.\ was a student researcher at Google Research.} \ \ \ \ \ Tal Linzen$^{5}$ \\
$^1$Northeastern University \ \ \ $^2$Technion -- Israel Institute of Technology \\
$^3$Google DeepMind \ \ \ $^4$New York University \ \ \ $^5$Google Research}

\begin{document}
\maketitle
\begin{abstract}
In-context learning (ICL) is now a common method for teaching large language models (LLMs) new tasks: given labeled examples in the input context, the LLM learns to perform the task without weight updates. Do models guided via ICL infer the underlying structure of the task defined by the context, or do they rely on superficial heuristics that only generalize to identically distributed examples? We address this question using transformations tasks and an NLI task that assess sensitivity to syntax---a requirement for robust language understanding. We further investigate whether out-of-distribution generalization can be improved via chain-of-thought prompting, where the model is provided with a sequence of intermediate computation steps that illustrate how the task ought to be performed. In experiments with models from the GPT, PaLM, and Llama~2 families, we find large variance across LMs. The variance is explained more by the composition of the pre-training corpus and supervision methods than by model size; in particular, models pre-trained on code generalize better, and benefit more from chain-of-thought prompting.
\end{abstract}

\section{Introduction}
Language models (LMs) have become increasingly important subjects of study due to their expressive power and performance at scale. When training large language models (LLMs) on massive amounts of text, surprisingly sophisticated linguistic behaviors, such as in-context learning (ICL), emerge \citep{brown-2020-gpt3,min-etal-2022-metaicl}: given only a small number of labeled training examples in the input context, LLMs can generalize to new instances of the task without weight updates. Thus, even without access to the model's weights, we can teach an LLM to perform new tasks with significantly higher-than-chance performance. This raises questions as to whether context is sufficient for LLMs to learn the underlying structure of a task, as opposed to superficial heuristics that do not generalize well. Indeed, LLMs have demonstrated counterintuitive biases in ICL settings \citep{pan2023incontext,min-etal-2022-rethinking,webson-pavlick-2022-prompt}, giving reason for skepticism. In this study, we ask: How robust is ICL to distribution shifts between in-context exemplars and test examples?

\begin{figure}[t]
    \centering
    \includegraphics[width=\linewidth]{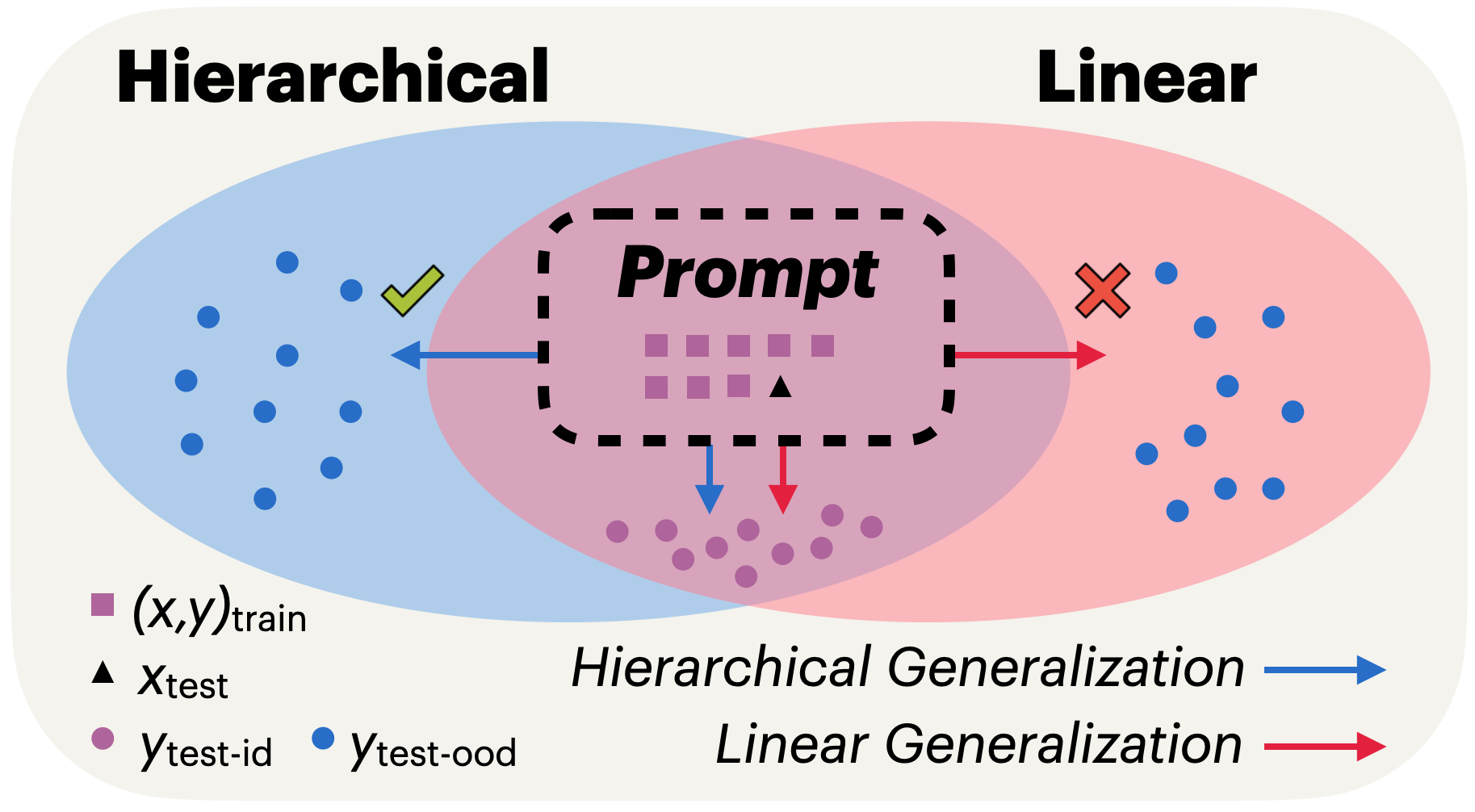}
    \caption{The syntactic transformations paradigm. We prompt language models with labeled examples $(x,y)_\text{train}$ that can be explained using either robust syntactic/hierarchical features or spurious positional/linear features. We also include the input from a test example $x_\text{test}$. We ensure the models have learned the task by evaluating on in-distribution examples $(x,y)_\text{test-id}$. Then, we observe whether models generalize syntactically or linearly on out-of-distribution examples $(x,y)_\text{test-ood}$.}
    \label{fig:transformations_summary}
\end{figure}

We investigate these questions using the test case of syntactic generalization. Accurate syntactic representations are necessary for robust language understanding. In LMs, syntax acquisition is causally associated with significant and abrupt drops in loss and improved performance on NLP tasks \citep{chen2024sudden}. Language is structured hierarchically, but the structure of sentences is not provided to models as part of their input. Therefore, a model could incorrectly assume that sentences have \emph{linear} rather than \emph{hierarchical} structure---e.g., that syntactic dependencies consist of adjacent words---and thus give an incorrect or unrelated answer to questions such as the following:

\ex. Is it true that \textbf{\blue{children}} who don't like \textbf{\red{sweets}} \textbf{\textcolor{violet}{are}} rare?

The \textbf{\blue{main subject}} is ``\textbf{\blue{children}}'', but a model that assumes sentences are structured linearly could instead analyze ``\textbf{\red{sweets}}'' as the subject, as it is the positionally closest noun to the \textbf{\textcolor{violet}{main verb}}. In downstream tasks, such assumptions about sentence structure would lead to incorrect responses to inputs that humans find simple \cite{mccoy-etal-2019-right}. Studies using syntactic transformations tasks \citep{frank-mathis-2007-transformational} have found that pre-training on large corpora imparts syntactic inductive biases to LMs \citep{mueller-etal-2022-coloring}, and the strength of syntactic preferences increases with model depth \citep{mueller-linzen-2023-plant}. These studies relied on fine-tuning, where the LM's weights are updated based on examples of the task; we extend these analyses to the ICL setup.

We investigate LLMs' generalization in syntactic transformations and natural language inference tasks. In each task, we evaluate in two settings: one where models are provided only with a handful of instances of the task, and another in which they are additionally provided with chain-of-thought (CoT) traces \citep{wei-etal-2022-cot}. CoT refers to the finding that LMs' task performance can be improved if they are instructed to predict the intermediate computations required for the task.
We examine whether giving LMs access to syntactic reasoning traces and meta-linguistic information increases their reliance on robust features.
We evaluate not only the accuracy of the final answer, but also the \emph{correctness} of the model's reasoning and the \emph{faithfulness} of the model's final answer to its own reasoning.

We find that, while all models perform well on in-distribution examples, even very large LMs are prone to relying on surface heuristics. 
Further, chain-of-thought results can be misleading: CoT improves in-distribution performance, but often \emph{decreases} out-of-distribution performance. This underscores the importance of out-of-distribution evaluation. Finally, we present evidence that exposure to code during pre-training assists models in overcoming these limitations.\footnote{Our code and data are available at \url{https://github.com/aaronmueller/syntax-icl}.}

\section{Experimental Setup}

We first examine syntactic transformation tasks, where the training and test instances are drawn from distributions that are distinct in controlled ways (\S\ref{sec:syntactic_transformations}). We test the syntactic generalization of multiple families of LMs (\S\ref{sec:models}) when prompted using in-context learning (\S\ref{sec:prompts}).

\subsection{Syntactic Transformations}
\setlength{\Exlabelwidth}{0.9em}
\setlength{\SubExleftmargin}{1.5em}

\label{sec:syntactic_transformations}

\paragraph{Question formation.} Here, a model is given a declarative sentence and must transform it into a yes/no question by moving the main auxiliary verb to the start of the sentence. For instance, given this training example:

\ex. The \textbf{yaks} near my salamanders \textbf{have} amused your unicorn.\label{ex:transformation_ambiguous}

\begin{table*}
    \centering
    \scriptsize
    \begin{tabularx}{\linewidth}{lrrrX}
        \toprule
        \textbf{Model} & \textbf{Params} (est.) & \textbf{Pre-train Tokens} (est.) & \textbf{\% Code} (est.) & \textbf{Additional Training} \\
        \midrule
        GPT-3 \texttt{text-davinci-001} & 175B & 400B & ? & Fine-tuned on human demonstrations and highly rated model outputs \\\grayline
        GPT-3.5 \texttt{code-davinci-002} & ? & 500B & 20\% & -- \\\grayline
        GPT-3.5 (Turbo) \texttt{gpt-3.5-turbo-0301} & ? & 500B+ & ? & ? \\\grayline
        GPT-3.5 \texttt{text-davinci-002} & ? & 500B+ & ? & Fine-tuned on human demonstrations and highly rated model outputs \\\grayline
        GPT-3.5 \texttt{text-davinci-003} & ? & 500B+ & ? & Reinforcement learning on human feedback \\\grayline
        GPT-4 \texttt{gpt-4-0314} & ? & ? & ? & ? \\
        \midrule
        PaLM & 540B & 780B & 5\% & -- \\\grayline
        Flan-PaLM & 540B & 782B & 5\% & Fine-tuned on human demonstrations \\
        \midrule
        Llama~2 & 70B & 2T & 4\% & --  \\\grayline
        CodeLlama & 34B & 2.5T & 20\% & Fine-tuned on human and model-generated\newline demonstrations\\
        \bottomrule
    \end{tabularx}
    \caption{Models used in this study, their estimated number of parameters and pre-training tokens, and the proportion of the pre-training corpus estimated to be source code. We use the largest sizes of each PaLM and (Code)Llama model, as larger models are typically better able to leverage in-context guidance \citep{wei2022finetuned}.}
    \label{tab:models}
\end{table*}

The model should move the main auxiliary verb ``have'' to the start of the sentence to form the question, ``\textbf{Have} the yaks near my salamanders amused your unicorn?''. The model should rely on hierarchical syntactic information $s$: it should detect the \emph{main auxiliary} in the sentence and move it to the front (the \textsc{Move-main} hypothesis). However, the model could instead learn the positional heuristic $p$ and move the \emph{linearly first} auxiliary in the sentence (the \textsc{Move-first} hypothesis). Both $s$ and $p$ produce correct outputs for the training examples, but only $s$ generalizes correctly to out-of-distribution inputs where, crucially, the main auxiliary is \emph{not} linearly first in the sentence: 

\ex. My \textbf{\blue{zebras}} that \textbf{\red{have}} admired the newt \textbf{\blue{haven't}} observed the peacocks. \label{ex:transformation_main}
\a. $^\drsh$\textsc{Move-main}: \textbf{\blue{Haven't}} my \textbf{\blue{zebras}} that \textbf{\red{have}} admired the newt observed the peacocks?
\b. $^\drsh$\textsc{Move-first}: *\textbf{\red{Have}} my \textbf{\blue{zebras}} that admired the newt \textbf{\blue{haven't}} observed the peacocks?

\paragraph{Tense reinflection.} Here, the task is to convert past-tense verbs into present-tense verbs. We want the model to detect the subject of each verb and reinflect it based on its \emph{subject}'s number $s$ (the \mbox{\textsc{Agree-subject}} hypothesis). Here, the training distribution consists of examples where all nouns have the same grammatical number:

\ex. Her \textbf{newt} around your \textbf{unicorn} \textbf{confused} some quail.\label{ex:tense_ambiguous}

Because all nouns have the same number, a model could learn to correctly convert ``\textbf{confused}'' to ``\textbf{confuses}'' by simply agreeing the verbs with the \emph{closest} noun $p$ (the \textsc{Agree-recent} hypothesis). To evaluate which hypothesis the model has learned, we evaluate on examples where only \textsc{Agree-subject} produces correct outputs:

\ex. The \textbf{\blue{yak}} upon my \textbf{\red{ravens}} \textbf{\blue{entertained}} her zebras.\label{ex:tense_subject}
\a. $^\drsh$\textsc{Agree-subject}: The \textbf{\blue{yak}} upon my \textbf{\red{ravens}} \textbf{\blue{entertains}} her zebras.
\b. $^\drsh$\textsc{Agree-recent}: *The \textbf{\blue{yak}} upon my \textbf{\red{ravens}} \textbf{\red{entertain}} her zebras.

\textbf{Task formulation.} Each syntactic transformation example ($x,y$) consists of input sentence $x$ and output sentence $y$, where $x$ and $y$ are semantically and lexically nearly identical but differ in the syntactic arrangement of the words (as described above).

There is a training set $\mathcal{S}_\text{train}$ and two test sets: an in-distribution (ID) test set $\mathcal{S}_\text{test-id}$ used to determine whether the model has learned to perform the task, and one out-of-distribution (OOD) test set $\mathcal{S}_\text{test-ood}$ used to determine whether the model generalizes in a manner consistent with the latent hierarchical structure of language. The distributions of $\mathcal{S}_\text{train}$ and $\mathcal{S}_\text{test-ood}$ differ in controlled ways: the training examples $\mathcal{S}_\text{train}$ could be correctly transformed using either syntactic feature $s$ or positional feature $p$, whereas the test examples \emph{require} reliance on $s$ for correct answers. In other words:
\begin{equation*}
\begin{aligned}
    \mathcal{S}_{\text{train}} &= \{(x,y) \mid s(x)=1 \wedge p(x)=1\}\\
    \mathcal{S}_{\text{test-id}} &= \{(x,y) \mid s(x)=1 \wedge p(x)=1\}\\
    \mathcal{S}_{\text{test-ood}} &= 
    \{(x,y) \mid s(x)=1 \wedge p(x)=0\}
\end{aligned}
\end{equation*}

See Figure~\ref{fig:transformations_summary} for an illustration. We prompt models with up to 8 exemplars $\{(x_1,y_1)_\text{train},\ldots,(x_8,y_8)_\text{train}\}$, followed by the input of an example from one of the test sets $x_\text{test}$. A model that has learned to rely only on the spurious feature $p$ will obtain 0\% accuracy on $\mathcal{S}_\text{test-ood}$, but 100\% on $\mathcal{S}_\text{train}$ and $\mathcal{S}_\text{test-id}$; only reliance on the syntactic feature $s$ will yield 100\% accuracy on $\mathcal{S}_\text{test-ood}$.

\subsection{Models}
\label{sec:models}

We use a series of Transformer-based \citep{vaswani-2017-attention} decoder-only autoregressive language models that are known to support in-context learning. Estimates of model sizes and pre-training corpus sizes are presented in Table~\ref{tab:models}. Estimates are derived from OpenAI\footnote{The information OpenAI released about their models was found here: \url{https://platform.openai.com/docs/model-index-for-researchers}. This information has been taken down, but is largely replicated in \citet{ye2023comprehensive}.} and prior work, including \citet{kim-schuster-2023-entity} and \citet{ye2023comprehensive}. As OpenAI does not provide official parameter counts or training set descriptions,\ the true numbers for GPT models may differ.

\begin{figure*}[t]
    \centering
    \includegraphics[width=0.95\linewidth]{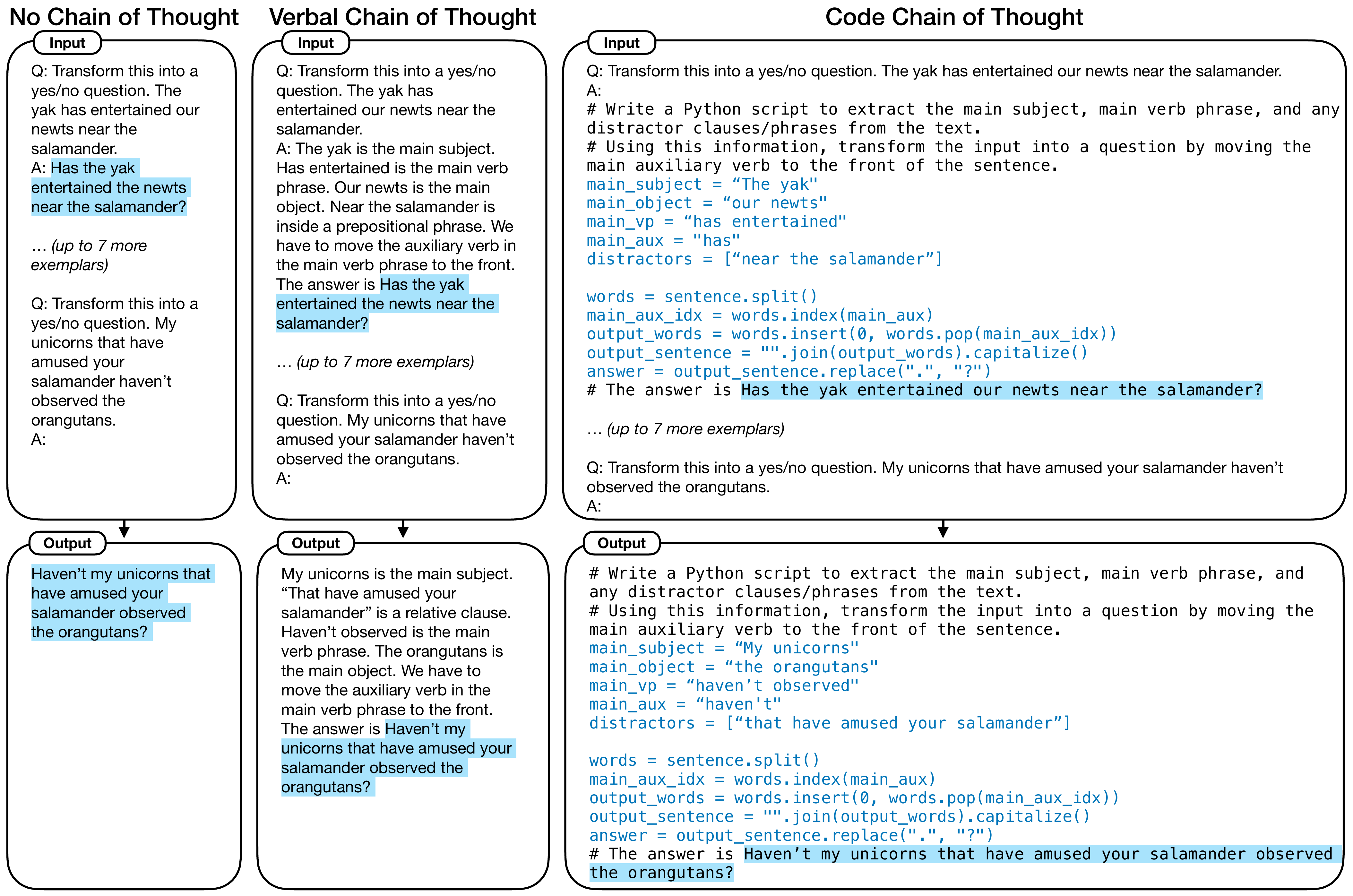}
    \caption{The prompt formats we use for question formation. We give the model up to 8 exemplars, followed by a test example. The model must generate the chain-of-thought reasoning \emph{and} the answer. We highlight the answers with a \hl{blue background}. In the code prompt, we put code (except comments) in \textcolor{bluetext}{\texttt{blue text}}. We construct similar prompts for tense reinflection; see Appendix~\ref{app:tense_reinflection_prompts}.}
    \label{fig:prompt_formats}
\end{figure*}

The OpenAI GPT models we use include GPT-3 \texttt{text-davinci-001} \citep{brown-2020-gpt3,ouyang-2022-instructgpt}, GPT-3.5,\footnote{\url{https://platform.openai.com/docs/models/gpt-3-5}} and GPT-4.\footnote{\url{https://platform.openai.com/docs/models/gpt-4}} The \mbox{GPT-3.5} variants we use include \texttt{code-davinci-002} \citep{chen-2021-codex}, which is pre-trained on natural language and a large amount of source code; \texttt{text-davinci-002}, a \texttt{code-davinci-002} model which is fine-tuned on instructions and human-labeled examples from many tasks \citep{wei2022finetuned}; and \texttt{text-davinci-003}, which is pre-trained similarly to \texttt{text-davinci-002} and then trained via reinforcement learning on human feedback (RLHF; \citealp{ouyang-2022-instructgpt}). GPT-3.5 (Turbo) builds on \texttt{text-davinci-003}'s methods, optimizing it for chat. Each of these models is estimated to be pre-trained on 400B or more tokens of text, as each is based on the 300B tokens of \citet{brown-2020-gpt3} plus 100B for instruction tuning as described in \citet{ouyang-2022-instructgpt}; models based on \texttt{code-davinci-002} are additionally estimated to be pre-trained on at least 100B tokens of code \citep{chen-2021-codex}. Few details of GPT-4's architecture or pre-training have been made public. 

PaLM models \citep{chowdhery-2023-palm} are pre-trained on large amounts of English text and code; 5\% of the pre-training corpus for PaLM is source code. The variants we test have 540B parameters. Flan-PaLM \citep{chung-2022-scaling} is further trained using instruction-finetuning \citep{wei2022finetuned}. 

Finally, we evaluate Llama~2 models \citep{touvron-2023-llama}, which are among the few open-source LMs capable of in-context learning. We use the 70B-parameter model,\footnote{For Llama~2 (70B), we use 4-bit quantization so that the model will fit on a single GPU.} which is trained on 2T tokens of mostly natural language and an unspecified amount of code. We also use CodeLlama \citep{roziere-2023-code}, which is Llama~2 trained on an additional 500B tokens of code. In preliminary experiments, we found that the 34B-parameter instruction-tuned variant of CodeLlama\footnote{\texttt{codellama/CodeLlama-34b-Instruct-hf}}
performed best, so we use that variant here.

For OpenAI and PaLM models, we use greedy decoding and set the temperature to 0 to reduce variance. For Llama models, we use settings from \citet{touvron-2023-llama}: nucleus sampling with \mbox{$p=0.9$} and temperature $0.1$.\footnote{We use different decoding hyperparameters because replicability is more crucial for models that are more difficult and costly to access. For Llama models, we consider slightly higher variance to be more acceptable in exchange for higher expected performance, for replication and trying other settings are both more accessible.}

\subsection{Prompt Formats}\label{sec:prompts}
In ICL, a set of labeled examples, concatenated into a ``prompt'', is provided in the model's context. We use 8 labeled examples (henceforth, \emph{exemplars}).\footnote{We attempted zero-shot evaluations to avoid confounds from the exemplar set. These did not score well: the models would typically drop the auxiliaries and change the past participle into simple past form; for example, ``The newts saw the yak.'' Or they would simply generate labels that we did not prompt for, such as ``True'' or ``No''.} Question formation exemplars are drawn from the training portion of \citet{mueller-etal-2022-coloring}. Tense reinflection exemplars are drawn from the training portion of \citet{mccoy-2020-syntax-trees}.

\begin{figure*}
    \centering
    \includegraphics[width=0.98\linewidth]{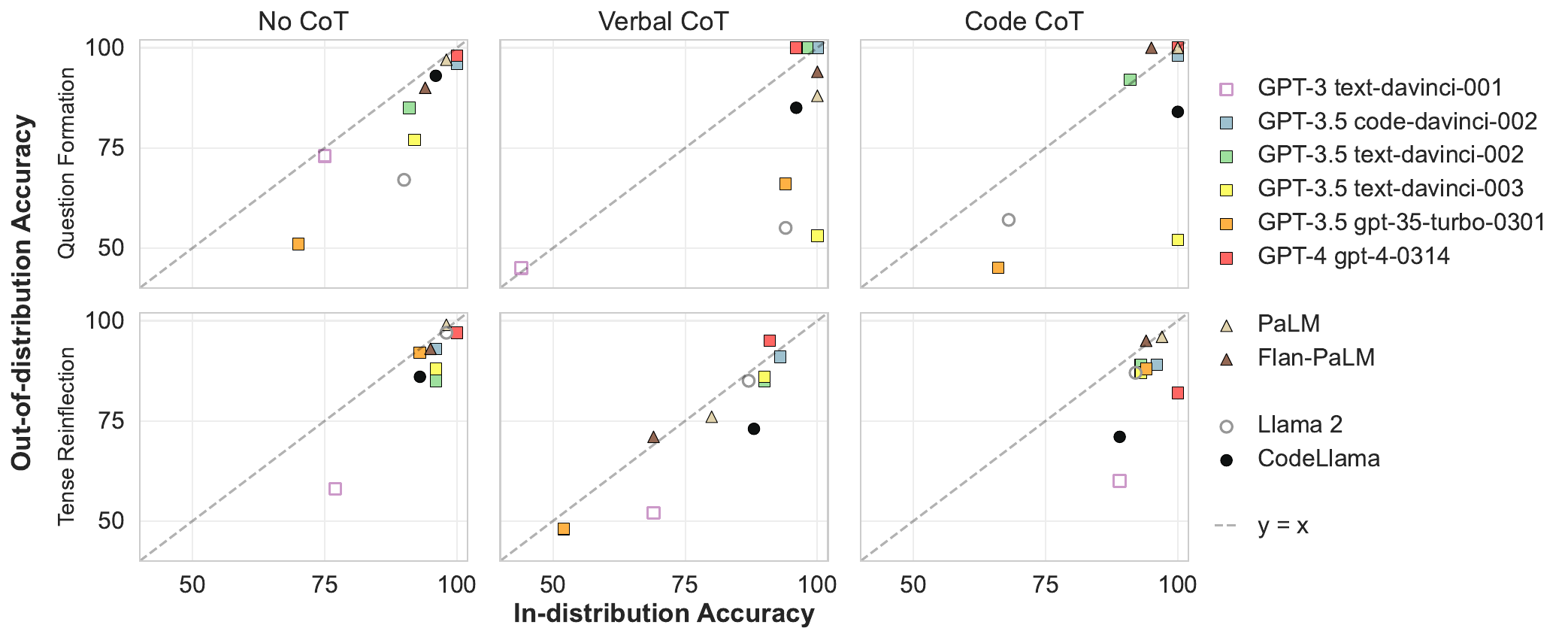}
    \caption{Main auxiliary accuracy on question formation and verb accuracy on tense reinflection. In-distribution accuracies reveal whether the models have learned the task, and out-of-distribution accuracies reveal whether models generalize robustly. Unfilled shapes (GPT-3, Llama~2) were trained on less than 5\% code. We interpret the dashed line as a ceiling on OOD accuracy given ID accuracy.}
    \label{fig:accs_transformations}
\end{figure*}

We also experiment with two types of manually written CoT traces, which explicitly provide cues to the syntactic structure of the exemplars; these are provided between the input sequence and the answer. The Verbal CoT prompt includes verbal descriptions of the syntactic structure of the sentence, where the model must describe which words compose the main subject, the main verb phrase, and other components. The Code CoT prompt likewise describes which words correspond to which syntactic components, but in a Python code format where each component is represented as a variable, and then the model is explicitly instructed (in code) how to combine these components to form the final answer. The format of the inputs is shown in Figure~\ref{fig:prompt_formats}. We search over prompt formats by tuning over training accuracy on 100 examples that were not part of the exemplar set.

For the Code CoT prompt, the 8 exemplars exceed the maximum sequence length of GPT-3 and (Flan-)PaLM; in such cases, we use 4 Code CoT exemplars, which is the maximum that will fit while still allowing the model to generate the full reasoning trace and answer.\footnote{We observe in our results that (Flan-)PaLM ID and OOD performance are still high for Code CoT despite using only 4 exemplars, and that GPT-3 does not achieve high ID or OOD performance given \emph{any} prompt (regardless of the number of exemplars).}

\subsection{Evaluation} 
We evaluate on a uniform subsample of 100 examples from the question formation generalization set of \citet{mueller-etal-2022-coloring} and 100 examples from the tense reinflection generalization set of \citet{mccoy-2020-syntax-trees}. To verify whether models have learned the task, we also present scores on 100-example uniform subsamples from the \emph{in-distribution} test sets, where each example can be correctly transformed using either syntactic \emph{or} positional features. All examples are generated from a probabilistic context-free grammar; as such, the distribution of syntactic structures is highly constrained, and the variation across examples is primarily lexical. We therefore reason that 100 examples should provide a representative sample of the syntactic structures in the evaluation sets.

To evaluate question formation examples, we use \textbf{main auxiliary accuracy}, which measures whether the model has moved the correct main auxiliary to the front of the sentence. This is measured by observing whether the first word is correct. For tense reinflection, we use \textbf{verb accuracy}, which measures whether each generated verb is correctly inflected \emph{and} in the correct relative position.\looseness=-1

\section{Results}\label{sec:results}

All models except GPT-3 and GPT-3.5 (Turbo) learn to perform the transformations tasks, as indicated by high in-distribution accuracies. Among models that learn the task well, there is large variance with respect to out-of-distribution accuracies.

\textbf{Scale does not fully explain performance.} GPT-3, GPT-3.5 \texttt{text-davinci-002} and GPT-3.5 \texttt{text-davinci-003} are estimated to be the same size, but the GPT-3.5 models generalize better. Llama~2 and CodeLlama outperform GPT-3 and (Flan-)PaLM despite being far smaller, and CodeLlama significantly outperforms Llama~2 on question formation despite being smaller. What explains differences between models, then?

\textbf{Pre-training on code improves OOD generalization.} On average, scores are higher given \emph{any} prompt for models pre-trained on code: \mbox{CodeLlama} significantly outperforms Llama~2 on question formation (and performs comparably on tense reinflection), while GPT-3.5 \texttt{code-davinci-002} outperforms all other GPT-3 and GPT-3.5 models. This agrees with and extends the finding of \citet{mueller-linzen-2023-plant} that the domain of the pre-training corpus significantly affects syntactic generalization. This also agrees with recent findings that code pre-training improves other fundamental linguistic abilities (e.g., entity tracking; \citealp{kim-schuster-2023-entity}).

Conversely, \textbf{RLHF may harm OOD generalization.} GPT-3.5 \texttt{text-davinci-003} performs at least as well as other GPT-3.5 models on in-distribution examples, but it generalizes consistently \emph{worse} than other models which are \emph{fine-tuned} on human demonstrations (including \texttt{text-davinci-002} and CodeLlama). Thus, RLHF is effective when testing on in-distribution test sets, but it may actively harm a model's ability to reason beyond its given exemplars. This finding is preliminary, however, and should be confirmed in future work via ablations over RLHF in a variety of architectures and model sizes.

\textbf{Chain-of-thought prompting has different impacts on in-distribution and out-of-distribution accuracies.} For question formation, CoT prompting tends to increase ID accuracy while not improving OOD generalization; this is especially apparent for verbal CoT---and, for models trained on code, the code CoT prompt as well. This reveals the importance of testing OOD generalization when designing CoT prompts: \textbf{when chain-of-thought prompting improves performance on ID examples, improvements will not necessarily generalize to OOD examples.} For tense reinflection, verbal CoT equally harms both ID and OOD accuracy relative to No CoT prompts; conversely, Code CoT maintains ID accuracy while harming OOD accuracy. This generalizes the above point: depending on the task and prompt format, \textbf{CoT can have different impacts on ID and OOD performance}.

We qualitatively analyze model errors to investigate why models achieve imperfect OOD generalization (App.~\ref{app:analysis}). We find that \texttt{code-davinci-002} is sensitive to variable names, and that most errors are because models move \emph{affirmative} verbs instead of syntactically correct verbs.

\section{Syntactic Generalization in Text Classification: The Case of Natural Language Inference}\label{sec:hans}

So far, our analyses indicate that LLMs leverage syntactic information to varying extents when taught a task via ICL, and that code pre-training improves OOD generalization. In this section, we test whether these trends extend to classification tasks. We leverage the Heuristic Analysis for NLI Systems (\textsc{Hans}; \citealp{mccoy-etal-2019-right}) dataset, which contains natural language inference (NLI) examples designed to diagnose reliance on syntactic heuristics. One typically trains the model on the training set of another NLI dataset, such as MNLI \citep{williams-etal-2018-broad} or RTE \citep{dagan-etal-2006-rte}, and then uses \textsc{Hans} as an evaluation set to measure whether models rely on syntactic heuristics. \citet{si2023prompting} evaluate RoBERTa and GPT-3 on \textsc{Hans} after training on MNLI; they find that both models perform similarly, suggesting that scale may not be enough to overcome syntactic heuristics. We extend this analysis to a wider variety of LLMs and to chain-of-thought prompting.

\textsc{Hans} diagnoses reliance on lexical overlap, subsequence, or constituent heuristics. Models that systematically rely on any of these heuristics are expected to obtain 100\% accuracy on examples where the label is ``entailment'' because the heuristics happen to make the correct prediction; however, models that rely on these heuristics will also obtain 0\% accuracy on examples where the label is ``non-entailment''. We average scores across heuristics here; see App.~\ref{app:hans} for full results.

\begin{table}[t]
    \centering
    \resizebox{\linewidth}{!}{
    \begin{tabular}{lrrrr}
    \toprule
    & \multicolumn{2}{c}{Entailment} & \multicolumn{2}{c}{Non-entailment} \\\cmidrule(lr){2-3}\cmidrule(lr){4-5}
    \textbf{Model} & No CoT & Verbal CoT & No CoT & Verbal CoT \\\midrule
    GPT-3 \texttt{text-davinci-001} & 56 & 99 & 82 & 2 \\
    GPT-3.5 \texttt{code-davinci-002} & 92 & 97 & 81 & 59 \\
    GPT-3.5 \texttt{text-davinci-002} & 72 & 97 & 91 & 57 \\
    GPT-3.5 \texttt{text-davinci-003} & 97 & 100 & 70 & 48 \\
    GPT-3.5 \texttt{gpt-3.5-turbo-0301} & 82 & 8 & 39 & 9 \\
    GPT-4 \texttt{gpt-4-0314} & 97 & 97 & 64 & 58 \\
    \midrule
    PaLM & 98 & 99 & 60 & 55 \\
    Flan-PaLM & 100 & 100 & 39 & 46 \\
    \midrule
    Llama~2 & 94 & 98 & 68 & 51 \\
    CodeLlama & 89 & 68 & 69 & 79 \\
    \bottomrule
    \end{tabular}}
    \caption{Accuracies on \textsc{Hans}. Scores are split by label, and then by prompt format; we aggregate across syntactic heuristics (full table in App.~\ref{app:hans}). High scores on entailment coupled with low scores on non-entailment signify that the model relies on the syntactic heuristic to predict the label. Chain-of-thought can increase reliance on heuristics: Compared to No CoT, Verbal CoT often demonstrates higher scores on entailment, but significantly lower scores on non-entailment.}
    \label{tab:hans-main}
\end{table}

We evaluate LLMs on HANS in the ICL setting. The 8 ICL exemplars are drawn from MNLI, and the test examples are from HANS. Our results are based on a uniform subsample of 100 examples per heuristic \emph{and} per label. We test across 3 syntactic heuristics, and there are 2 labels (entailment and non-entailment); thus, we have 600 test examples in total. We use No CoT prompts and Verbal CoT prompts similar to those we depict in Figure~\ref{fig:prompt_formats}; see App.~\ref{app:hans} for examples of our prompt formats.

Our results (Table~\ref{tab:accs_hans}) suggest that LLMs are susceptible to syntactic heuristics, but to a lesser extent than smaller-scale LMs fine-tuned on MNLI \citep{mccoy-etal-2019-right}. Most models achieve near 100\% accuracy on examples where the gold label is ``entailment'', as expected. However, on ``non-entailment'' examples, models perform much more poorly. Unlike in the transformation tasks, code pre-training does not lend a significant advantage.

We also observe that \textbf{chain-of-thought can make models significantly more prone to relying on syntactic heuristics.} With Verbal CoT, scores on entailment examples generally increase (except for GPT-3.5 (Turbo)). On non-entailment examples, however, scores reduce to near-random-chance---and, for some models, to near-zero. This pattern suggests that models rely more extensively on the heuristic given Verbal CoT prompts. This reinforces the importance of detailed evaluation when tuning one's prompts: results from prompt tuning experiments can often be misleading if one only observes in-distribution accuracies, or overall (as opposed to label-specific) accuracies.

\section{Do LLMs Generalize Faithfully?}
LLMs have been found to generate answers which are not \textbf{faithful} to their own chain-of-thought reasoning \citep{lyu-etal-2023-faithful,turpin2023language}. In this section, we test the relationship between models' CoT reasoning and their answers. There are at least two possibilities that could lead to low accuracy: First, the models could produce incorrect CoT reasoning traces, but answer consistently with those traces. Second, they could produce correct reasoning traces, but ignore those traces when producing the output. Here, we evaluate the accuracy of models' reasoning, as well as the faithfulness of models' final answers to their reasoning traces.

\subsection{Method}
The templatic format of the Code CoT prompt allows us to easily extract each component of the model's reasoning from the variable values. We observe whether the reasoning matches what the ground-truth values should be (\textbf{reasoning accuracy}) and whether the model's prediction aligns with the generated reasoning (\textbf{faithfulness}).

For question formation, we measure how often the model extracts the \blue{main subject}, the \red{main verb phrase (VP)}, and the \violet{object}, and how often its output follows its analysis of each component:

\ex. \blue{The salamanders} near my yaks that haven't entertained your unicorn \red{have amused} \violet{the newt}. $\Rightarrow$ \red{Have} \blue{the salamanders} near my yaks that haven't entertained your unicorn \red{amused} \violet{the newt}?\label{ex:qf_reasoningacc}

\begin{figure*}[t]
    \centering
        \includegraphics[width=0.925\linewidth]{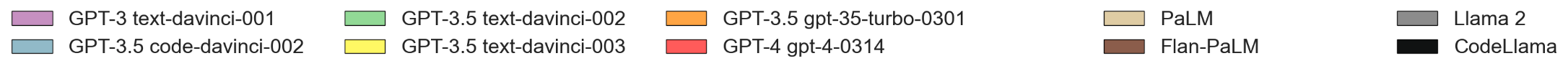}
        \vspace{0.05cm}
        
        \begin{subfigure}{0.49\linewidth}
            \centering
            \includegraphics[width=0.95\linewidth]{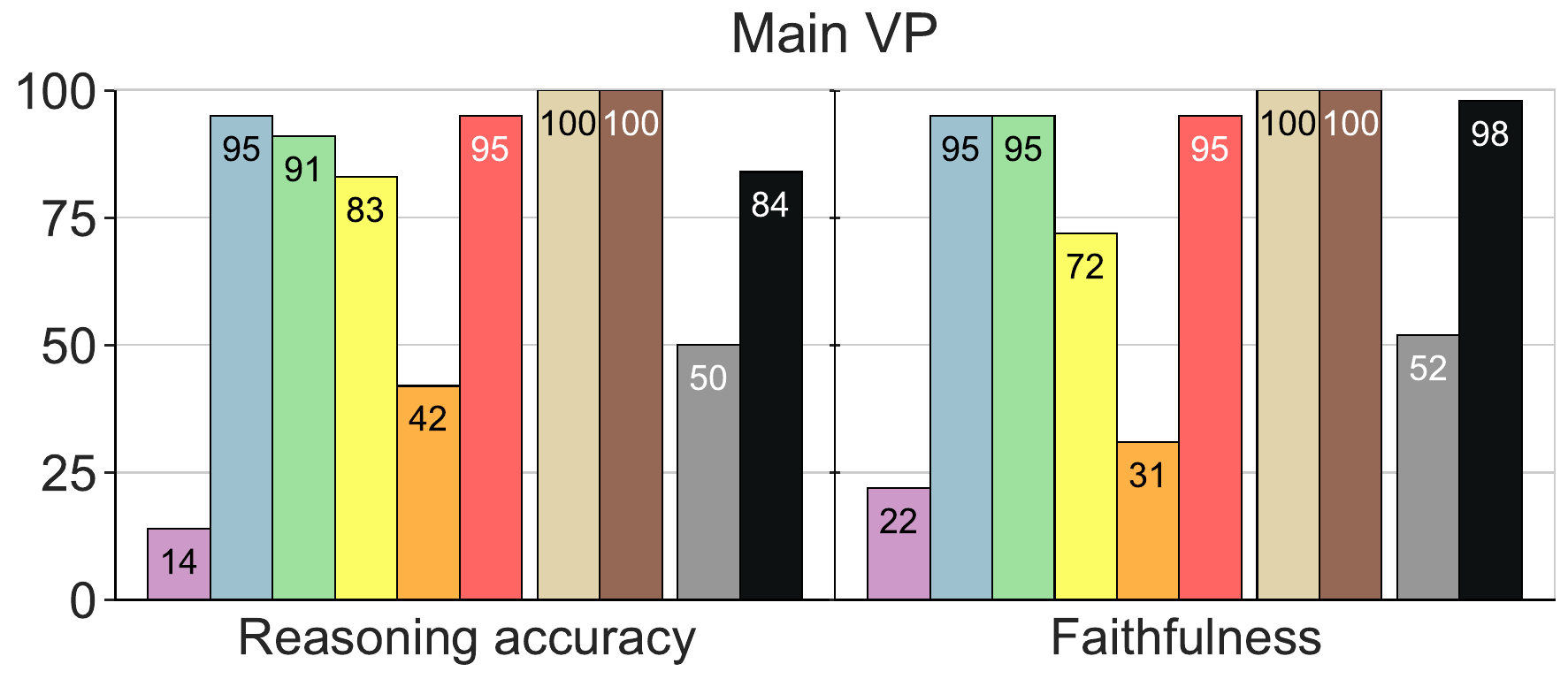}
            \caption{Question formation}
        \end{subfigure}
        \rulesep
        \begin{subfigure}{0.49\linewidth}
            \centering
            \includegraphics[width=0.95\linewidth]{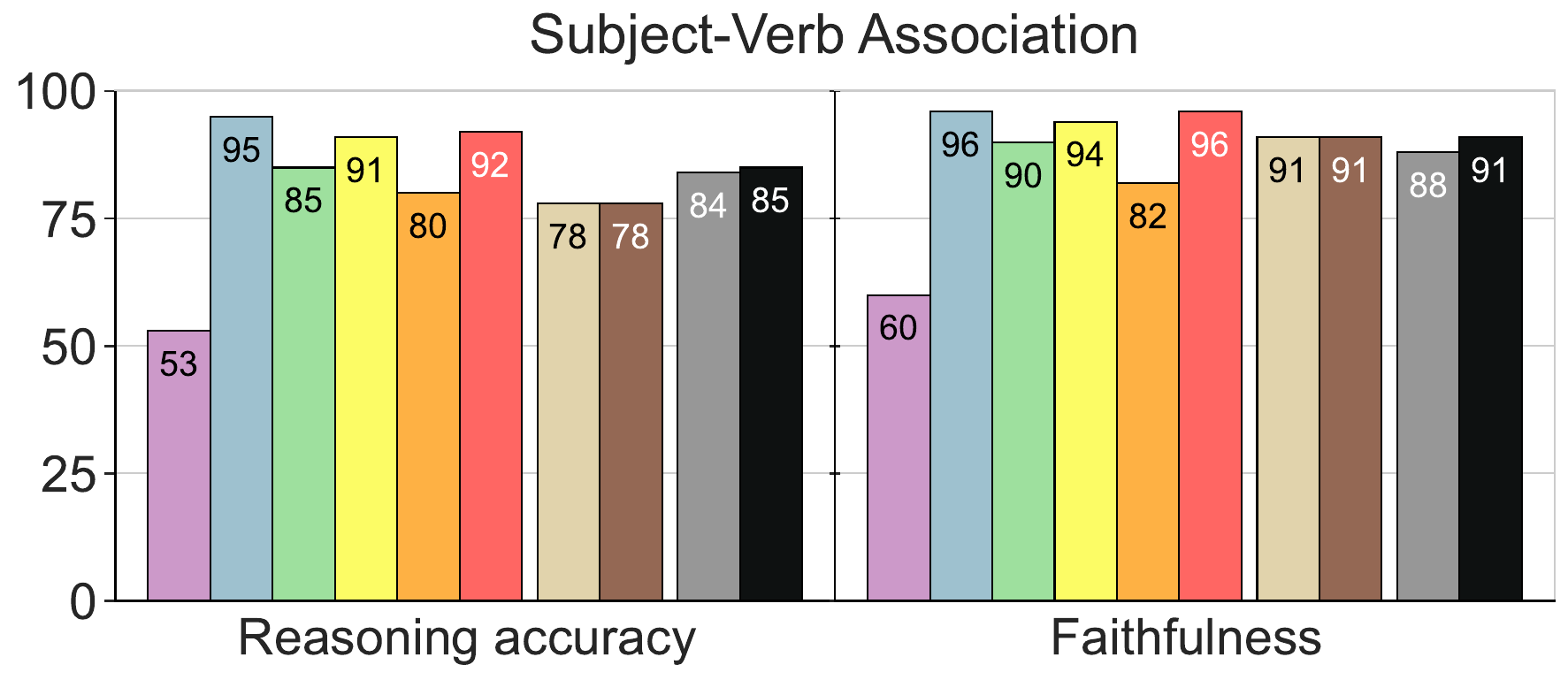}
            \caption{Tense reinflection}
        \end{subfigure}
\caption{Reasoning accuracies and faithfulness scores for question formation (left) and tense reinflection (right) using the Code CoT prompt. Reasoning accuracies and faithfulness are highest for \texttt{code-davinci-002}, GPT-4, (Flan-)PaLM, and CodeLlama.}
\label{fig:qf_tense_reasoningacc_faithfulness}
\end{figure*}

For tense reinflection reasoning, we measure whether each \blue{noun} is detected and whether each verb is correctly associated with its subject (\red{subject-verb association}). When evaluating faithfulness, we also measure whether the verbs in the output agree with the subjects they were associated with in the CoT reasoning (\violet{verb number}). For example, given the following prompt and answer:

\ex. Your \blue{peacocks} that \red{admired} the \blue{raven} \red{remembered} your \blue{vulture}. $\Rightarrow$ Your \blue{peacocks} that \violet{a}\red{d}\violet{m}\red{i}\violet{r}\red{e} the \blue{raven} \violet{r}\red{e}\violet{m}\red{e}\violet{m}\red{b}\violet{e}\red{r} your \blue{vulture}.\label{ex:tense_reasoningacc}\footnote{Alternating colors mean that a word is used in evaluating multiple components.}

We present a subset of our analysis, focusing on components that are essential for achieving the correct answer. 
See App.~\ref{app:faithfulness_full} for the full analysis and details on how we extract each component from a model's generated reasoning and answer.

\subsection{Results}

Reasoning accuracies and faithfulness (Figure~\ref{fig:qf_tense_reasoningacc_faithfulness}) are highest for \texttt{code-davinci-002}, GPT-4, \mbox{(Flan-)PaLM}, and CodeLlama. These are the same models which had the highest OOD accuracies; indeed, OOD accuracies correlate strongly with reasoning accuracies ($\rho_\text{Spearman}$~=~0.71, $p<$~.01) and faithfulness scores ($\rho_\text{Spearman}$~=~0.71, $p<$~.01). This is unsurprising, but it still aids our understanding of \emph{why} certain models generalize better. For question formation, models pre-trained on large amounts of code---notably, GPT-3.5 \texttt{code-davinci-002}, \texttt{text-davinci-002}, GPT-4, Flan-(PaLM), and CodeLlama---show high reasoning accuracies and faithfulness. GPT-3.5 \texttt{text-davinci-003} reasons less accurately, perhaps due to RLHF. Trends are similar for tense reinflection, except that Llama~2 and CodeLlama perform more similarly to each other and PaLM/Flan-PaLM perform worse than GPT-3.5 models. These results raise two hypotheses: (1) code pre-training induces better OOD generalization because it induces better (more accurate and faithful) reasoning, whereas (2) RLHF induces worse generalization because it optimizes features that are orthogonal to linguistic reasoning. These hypotheses are discussed in \S\ref{sec:discussion}.

We also observe that \textbf{reasoning accuracy correlates strongly with faithfulness} ($\rho_\text{Spearman}$~=~0.87, $p<$~.01). Our hypothesis was that low performance could be explained by models either reasoning well but ignoring their reasoning, or reasoning poorly and answering accordingly. Instead, we find that a model's ability to follow its own reasoning is linked to how well it reasons. Future work should investigate to what extent accuracy, reasoning accuracy, and faithfulness are \emph{causally} interdependent.

\section{Discussion}\label{sec:discussion}
LLMs guided via ICL often explain the structure of input examples and tasks using spurious positional and word-level features (\S\ref{sec:results}), as well as spurious syntactic heuristics (\S\ref{sec:hans}). It is surprising that LLMs generalize in a manner not consistent with English grammar, given that language models are directly optimized over a large quantity of long contexts to produce probable sequences. Larger and deeper models generally behave in a manner more consistent with syntactic structure when fine-tuned \citep{mueller-linzen-2023-plant}, but this trend does not generalize to ICL. Thus, despite impressive performance on downstream tasks, our findings provide evidence that \textbf{LLMs do not consistently leverage the latent structure of language when processing or generating language.} 
LLMs will therefore struggle to generalize well outside of their exemplars' distribution.

\textbf{Chain-of-thought has significantly different impacts on ID vs.\ OOD examples.} In transformations tasks, it sometimes improved ID performance while maintaining or decreasing OOD performance; on \textsc{Hans}, it significantly increased reliance on syntactic heuristics. This reveals an actionable takeaway: one should perform thorough evaluations when prompt tuning. Results from overall (rather than label-specific) accuracies and in-distribution examples can be misleading. A caveat is that one could always tune these prompts further to obtain better performance, just as one could always tune hyperparameters to obtain better models.

\textbf{Code pre-training improves LLM generalization.} Various studies observe that LMs trained on large amounts of code perform better on various NLP tasks and linguistic evaluations \citep{kim-schuster-2023-entity,madaan-etal-2022-language,sap-etal-2022-neural}. Why is code such an effective signal when mixed with natural language? Some have speculated that code provides additional grounding \citep[][\emph{inter alia}]{potts-2020-understand,merrill-etal-2021-provable}. Perhaps more importantly, code contains frequent instances of long-range state tracking, as well as hierarchically structured classes and function stacks; these may impart inductive biases that are helpful for learning hierarchical linguistic structure.

It is unclear why GPT-3.5 \texttt{text-davinci-003}---the only model in the GPT-3.5 family trained with RLHF---performs significantly worse than comparable GPT-3.5 models. This goes against conventional wisdom that RLHF generally improves performance on downstream NLP tasks \citep{ye2023comprehensive},\footnote{OpenAI also officially states that \texttt{text-davinci-003} is more capable than \texttt{text-davinci-002} in multiple locations on their website, including \href{https://help.openai.com/en/articles/6779149-how-do-text-davinci-002-and-text-davinci-003-differ}{here} and \href{https://beta.openai.com/playground}{here}.\label{footnote:official_rec}} but corroborates preliminary evidence that RLHF degrades certain aspects of performance, like certainty calibration.\footnote{\url{https://openai.com/research/gpt-4}} Further work is needed to fully understand RLHF's impact on generalization and whether these findings are causally linked. Perhaps the reinforcement learning procedure optimizes models to generate language that \emph{semantically} aligns with human quality judgments, but at the cost of causing the model to assign lower importance to structural features. Meanwhile, providing human feedback via fine-tuning may be less optimal for aligning outputs to human judgments, but better for preserving sensitivity to syntax.

Larger models often perform better on many NLP tasks. However, our findings reveal that \textbf{scale is not a panacea for robust generalization}: rather, other factors like training objectives, the type of pre-training data, and the supervision method(s) make a significant difference. This agrees with and extends findings from \citet{mueller-linzen-2023-plant} that the domain of the pre-training data significantly influences how models generalize.

\section{Related Work}
Since the discovery that LMs are capable of ICL, studies have explored ICL's limits (e.g., \citealt{akyurek-etal-2023-what,chan-etal-2022-incontext}). Analyses of ICL have revealed counterintuitive biases: for example, correct labels are not necessary for strong performance \citep{min-etal-2022-rethinking}, and models can perform very well even given misleading/irrelevant prompts \citep{webson-pavlick-2022-prompt} or flipped/semantically misleading label names \citep{wei2023larger,pan2023incontext}. In contemporaneous work, \citet{si2023measuring} analyze which semantic features (e.g., sentiment vs.\ topic) LLMs prefer by designing underspecified exemplars. Their method is similar to ours in that they use ambiguous exemplars and disambiguating test examples; however, we apply this approach to \emph{syntactic} processes to analyze the fundamental linguistic structural abilities of LLMs.

Any finite training set is consistent with multiple generalizations. Crucially, most benchmarking tasks rely on data where the training and test sets are drawn from the same distribution, which limits our understanding of how well models truly generalize \citep{linzen-2020-accelerate}. \citet{saparov-he-greedy-2023} find that LLMs are prone to relying on spurious correlations that hinder robust generalization, while \citet{drozdov-2023-compositional} find that a series of chain-of-thought prompts can yield more robust generalization.
With respect to syntax, LSTM- and Transformer-based encoder-decoder models trained from scratch on syntactic transformations \emph{do not} generalize in a syntax-sensitive manner \citep{mccoy-etal-2018-poverty,mccoy-2020-syntax-trees,petty-frank-2021-transformers}, but pre-trained encoder-decoder models \emph{do} generalize syntactically \cite{mueller-etal-2022-coloring}. Similar positive results have been reported for RoBERTa, but only after large-scale pre-training \citep{warstadt-bowman-2020-linguistic,warstadt-etal-2020-learning}.

\citet{hu-levy-2023-prompting} prompt LLMs on metalinguistic judgments; they find that prompting underestimates syntactic awareness compared to probability measurements. Our findings extend this conclusion: this issue is more pronounced when test examples are not identically structured to the exemplars. When fine-tuning, larger models generally have inductive biases that align more strongly with syntactic structure \citep{mueller-linzen-2023-plant}, so it is plausible that greater scale should lead to more syntax-sensitive behavior when using ICL. Nonetheless, this is not what we find. 

\section{Conclusions}
We have investigated how well LLMs generalize out-of-distribution on tasks requiring syntax for robust performance. Our findings reveal significant variance across models that is not fully explained by scale. Models trained on code are better at leveraging in-context examples to generalize more robustly, and at reasoning accurately and faithfully---even at smaller scales.

\section*{Acknowledgments}
We thank Jacob Eisenstein for helpful comments on a previous draft of this paper. We also thank Microsoft for supporting experimentation with OpenAI models via Azure under the Accelerate Foundation Models Research program. Aaron Mueller was supported by a National Science Foundation Graduate Research Fellowship (Grant \#1746891). This work was supported in part through the NYU IT High Performance Computing resources, services, and staff expertise. 

\section*{Limitations}
Using closed-source language models presents many scientific challenges. There are no public resources that contain exact information about the models we test, such as their parameter counts, training distributions, and corpus sizes, among other important details. We also cannot definitively rule out that the transformations data is contained in the pre-training corpus. When asking the GPT models directly if they have seen the dataset or its GitHub repository, they did not state that they have seen them. However, Bard (not used in this study) directly admits to having seen syntax-aware evaluation datasets like \textsc{Hans}.

OpenAI support for the Codex model was discontinued while this project was in progress. This reveals another set of challenges of working with closed-sourced models in scientific contexts: long-term replicability becomes difficult, and access is dependent on the support of non-accountable entities. While we were able to attain access again by applying to a research program, access is not guaranteed indefinitely.

Regarding our experimental design, it is difficult to disentangle whether model errors are due to sub-human syntactic reasoning abilities, faulty inductive biases, or simply poor ICL abilities (App.~\ref{app:analysis}). We partially controlled for this by evaluating on in-distribution transformations, which evaluates how well models can leverage ICL on in-distribution examples. However, the transformations task does not in itself tell us what specific \emph{mechanism} leads to incorrect results: perhaps LLMs do robustly represent the syntactic structure of input sentences, but preferentially rely on superficial features. Or, perhaps they do not robustly represent sentence structure, instead relying on a series of heuristics that allow certain models to approximate syntactic generalization in specific circumstances. Future work should employ mechanistic interpretability methods to uncover what, precisely, causes errors in transformations tasks.

\section*{Ethics Statement}
Regardless of how well large language models generalize or to what extent they demonstrate language understanding, they are still eminently capable of misuse and harm. LLMs excel at generating convincing arguments for false conclusions, or for generating false information and confidently presenting it as fact---for example, when generating fake news. Our findings may erroneously suggest to some readers that addressing these concerns can wait, as LMs are still incapable of robustly understanding sentence structure. We disagree with this takeaway: research that addresses the harms of LLMs should occur in parallel with research that investigates (and improves) their decision-making mechanisms.

\bibliography{anthology,custom}

\newpage

\appendix
\section{Tense Reinflection Prompt Formats}\label{app:tense_reinflection_prompts}
\begin{figure*}
    \centering
    \includegraphics[width=\linewidth]{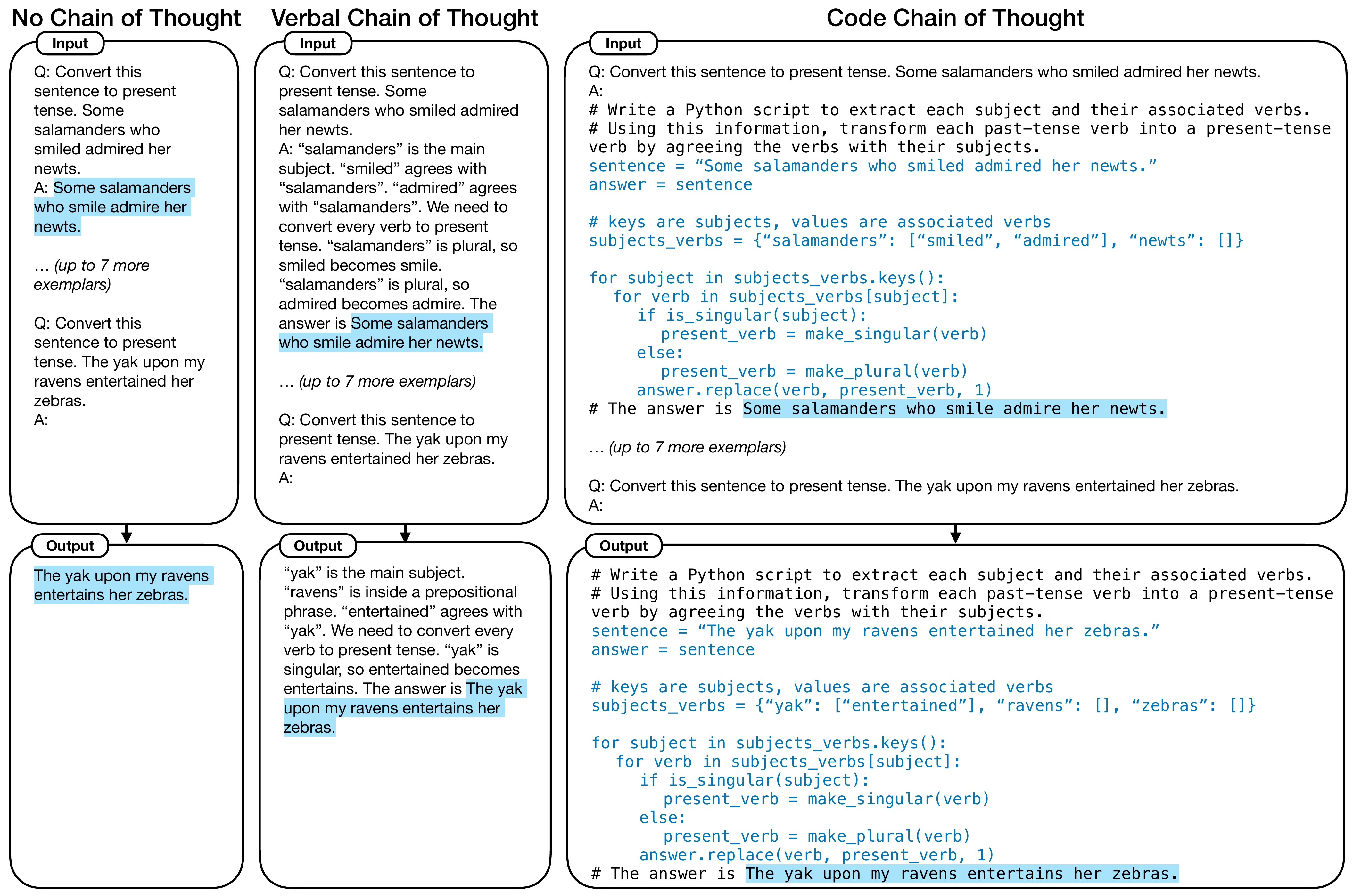}
    \caption{The prompt formats we use for tense reinflection. We give the model up to 8 exemplars, followed by a test example. The model must generate the chain-of-thought reasoning \emph{and} the answer. We highlight the answers with a \hl{blue background}. In the code prompt, we put code (except comments) in \textcolor{bluetext}{\texttt{blue text}}.}
    \label{fig:tense_reinflection_formats}
\end{figure*}

In Figure~\ref{fig:tense_reinflection_formats}, we present the prompt formats we use for the tense reinflection task.

\begin{figure*}
    \centering
    \includegraphics[width=\linewidth]{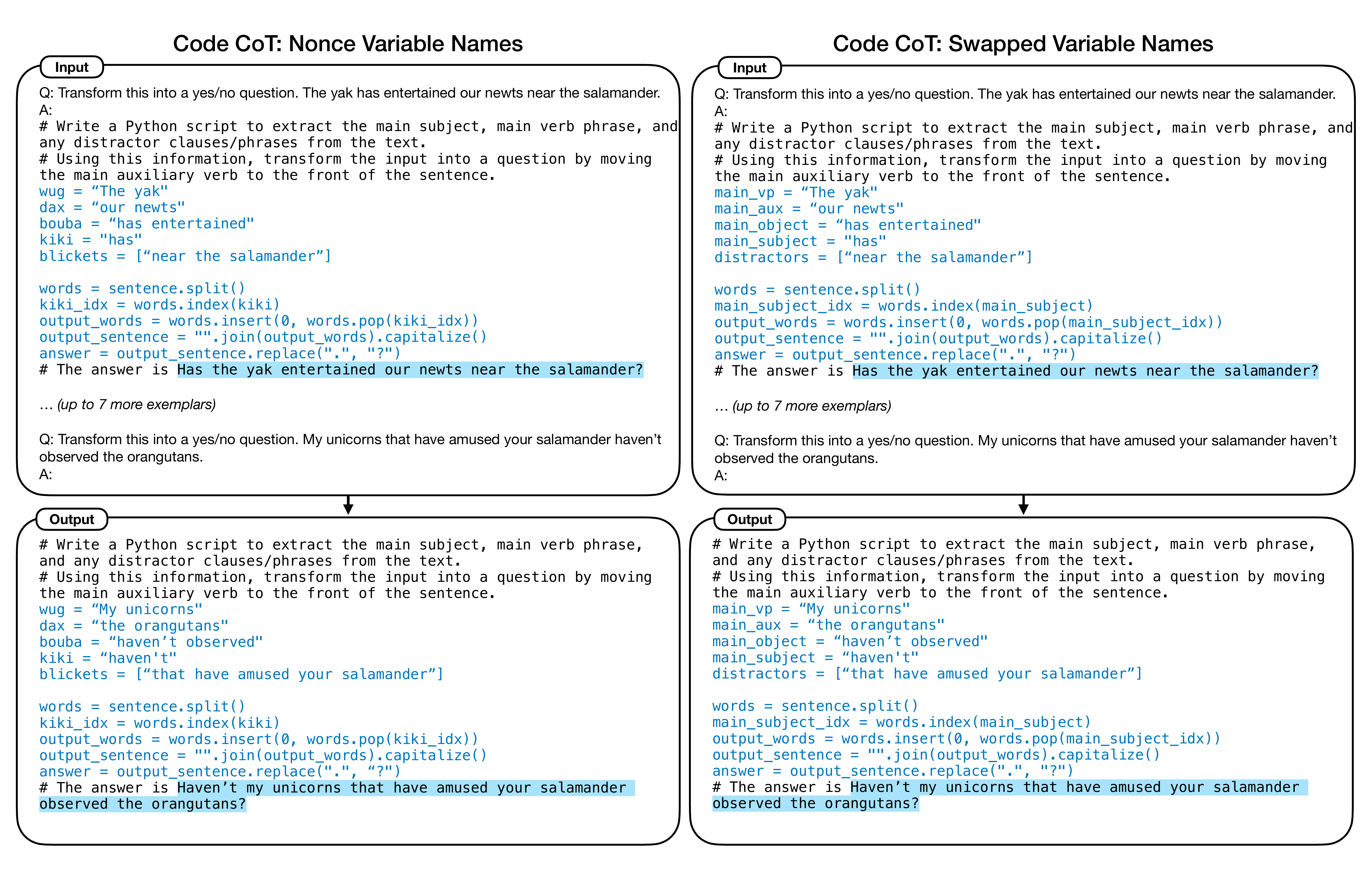}
    \caption{Prompt formats for our Code CoT analyses. For each exemplar, we ensure that the same variable name is used for the same syntactic component, such that the model can form meaningful associations with the nonce/swapped variable names.}
    \label{fig:code_ablation_formats}
\end{figure*}

\section{Ablations and Error Analysis}\label{app:analysis}
In this appendix, we analyze which aspects of the prompt affect OOD generalization and investigate the type of errors that LMs make.

\paragraph{When reasoning with code, GPT-3.5 \texttt{code-davinci-002} is sensitive to variable names, but GPT-4 is not.} GPT-4 and \texttt{code-davinci-002} perform best on both transformations, and seem to improve with Code CoT prompting. We note that in the prompts we used in the main text, the variables had semantically meaningful names, such as \texttt{main\_subject}; here, we ask how crucial this factor is for the models' ability to benefit from the Code CoT prompt. We present the prompt formats we use for this analysis in Figure~\ref{fig:code_ablation_formats}. 

We first replace variable names (like \texttt{main\_auxiliary} and \texttt{main\_subject}) with nonce words (like \texttt{wug} and \texttt{dax}). We ensure that the same variable name is used for the same syntactic component in each exemplar, such that the model can form meaningful associations with the nonce (or swapped, as in the following analysis) variable names. For \texttt{code-davinci-002}, this reduces 8-shot performance from 97\% to 76\%. For GPT-4, this only reduces 8-shot performance from 97\% to 96\%. 

In a second experiment, we assign variable names adversarially, such that seemingly meaningful variable names are no longer associated with their conventional linguistic category; for example, the main verb phrase is associated with the variable \texttt{main\_object} (e.g., \texttt{main\_object = "has entertained"}), or the main subject is associated with the variable \texttt{main\_vp} (e.g., \texttt{main\_vp = "The newt"}). For \texttt{code-davinci-002}, this reduces performance from 98\% to 83\%, but for GPT-4, the accuracy stays the same at 97\%. These patterns indicate that \texttt{code-davinci-002} is sensitive to---and may usably understand and rely on---terminology like ``subject'' and ``verb''. Meanwhile, GPT-4 may or may not understand the ``subject'' and ``verb'' names, but it does not crucially rely on them to perform the task; it instead shows an ability to adapt to arbitrary variable names.

\paragraph{GPT-3 and GPT-3.5 (Turbo) rely heavily on spurious lexical features.} What kinds of errors are causing such low performance in GPT-3 and \mbox{GPT-3.5 (Turbo)}? In a qualitative analysis, we find that when these models do not produce the correct answer, it is typically because they move the \emph{affirmative} (non-negated) auxiliary, regardless of whether it is the main auxiliary. We call this the \textsc{Move-affirmative} heuristic. For example, in Ex.~\ref{ex:affirmative_correct} and \ref{ex:affirmative_incorrect} below, these models would likely move the affirmative auxiliary, even though it results in the incorrect output in Ex.~\ref{ex:affirmative_incorrect}.

\ex. \a. The unicorn that \red{\textbf{hasn't}} entertained the newts \blue{\textbf{has}} observed the yak. $\Rightarrow$ \blue{\textbf{Has}} the unicorn that \red{\textbf{hasn't}} entertained the newts observed the yak? \raisebox{-.15\height}{\includegraphics[height=0.35cm]{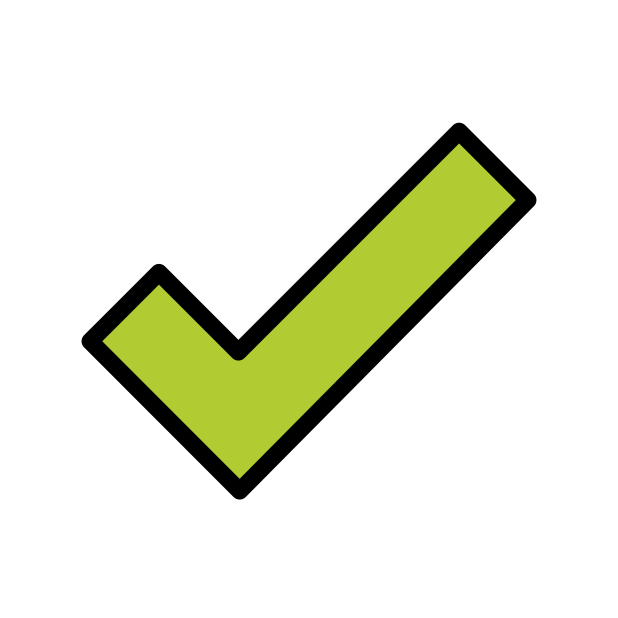}} \label{ex:affirmative_correct} 
\vspace{5pt}\b. The unicorn that \red{\textbf{has}} entertained the newts \blue{\textbf{hasn't}} observed the yak. $\Rightarrow$ \red{\textbf{Has}} the unicorn that \red{\textbf{has}} entertained the newts observed the yak? \raisebox{-.15\height}{\includegraphics[height=0.35cm]{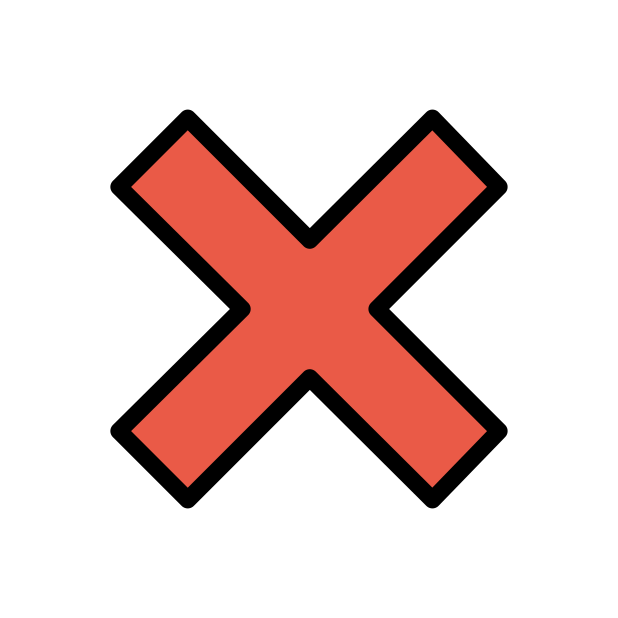}} \label{ex:affirmative_incorrect}

In fact, for \emph{all} models we test, errors in question formation are typically due to reliance on \mbox{\textsc{Move-affirmative}}.
The fact that models adopt this heuristic is puzzling, as it is not consistent with the in-context exemplars (half of the verbs in the exemplars are negative verbs), and therefore reflects an inductive bias. We hypothesize that the bias in question favors generating linguistically unmarked forms in ambiguous contexts. Such an unmarkedness bias would be easy to learn given that unmarked forms are generally more frequent \citep{greenberg1966language}. In this case, affirmative verb forms are likely more frequent than negative ones, especially at the beginning of a question. \footnote{The relationship between markedness and frequency is disputed \citep{battistella1996logic}. Frequency and markedness often correlate, but not always; see \citet{kucera-1982-markedness} for an empirical case study in English.} More broadly, a systematic preference for linguistically unmarked forms could be a form of overfitting to the training data. Future work could assess whether this type of overfitting is more likely with scale, or whether it could be overcome with some form of intervention to the data or model.

In tense reinflection, most errors are due to models relying on the linear \textsc{Agree-recent} hypothesis. Thus, errors in this case reflect reliance on word position and relative word ordering in generating sentences. All models we test generalize in a syntax-sensitive manner more often than not, but this error pattern reflects that models are not \emph{entirely} dependent on syntactic inductive biases: rather, given in-context examples, models can generalize using either inductive bias.

\section{Faithfulness of Chain-of-Thought Reasoning Traces: Full Analysis}\label{app:faithfulness_full}
\subsection{Extracting Components from Model Reasoning}
\paragraph{Question formation.} The model must generate variables corresponding to the main subject, main object, and main verb phrase. For example, given Ex.~\ref{ex:qf_reasoningacc}, the model must predict \texttt{\blue{main\_subject = "The salamanders"}} for its reasoning to count as correctly identifying the main subject. For faithfulness, if the model predicts \texttt{\blue{main\_subject = "The salamanders"}} in its reasoning and then begins its answer with ``Have the salamanders\ldots'', this would be considered faithful to its own reasoning on the main subject; conversely, if it predicts \texttt{\blue{main\_subject = "The salamanders"}} in its reasoning but then begins its final answer with ``Have the yaks\ldots'', this would not be considered faithful.

\paragraph{Tense reinflection.} The model must generate a Python dictionary of subjects (keys) and verbs (values) as part of its CoT prompt. In this example, the correct dictionary is \texttt{subjects\_verbs = \{\blue{"}\red{p}\blue{e}\red{a}\blue{c}\red{o}\blue{c}\red{k}\blue{s}\red{"}: \red{["admired", "remembered"]}, \blue{"}\red{r}\blue{a}\red{v}\blue{e}\red{n}\blue{"}: \red{[]}, \blue{"}\red{v}\blue{u}\red{l}\blue{t}\red{u}\blue{r}\red{e}\blue{"}: \red{[]}\}}. The \blue{noun} reasoning accuracy is the proportion of nouns from the sentence that are present in the dictionary, while the \red{subject-verb association} accuracy is the proportion of verbs associated with the correct subject key. For faithfulness, we measure how many \blue{nouns} from the dictionary are present in the answer, and whether the \red{verbs' subjects} in the answer are the same as they are in the dictionary (i.e., whether the sentence structure is compatible with the reasoning). For \violet{verb number} faithfulness, we measure the proportion of verbs in the dictionary whose grammatical number in the final answer is the same as the grammatical number of their subject in the dictionary.

\subsection{Full Results}
\begin{figure*}[t]
    \centering
    \includegraphics[width=0.9\linewidth]{syntax_icl/figures/faithfulness/legend.png}
    \vspace{0.05cm}
    
    \includegraphics[trim={0 0.95cm 0 2.9cm},clip,width=0.9\linewidth]{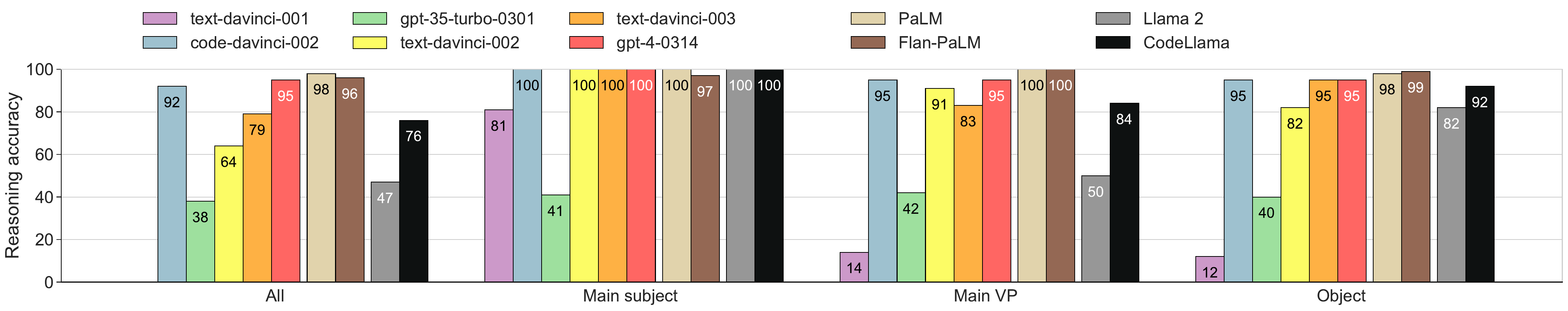}
    
    \vspace{0.3cm}
    
    \includegraphics[trim={0 0 0 2.9cm},clip,width=0.9\linewidth]{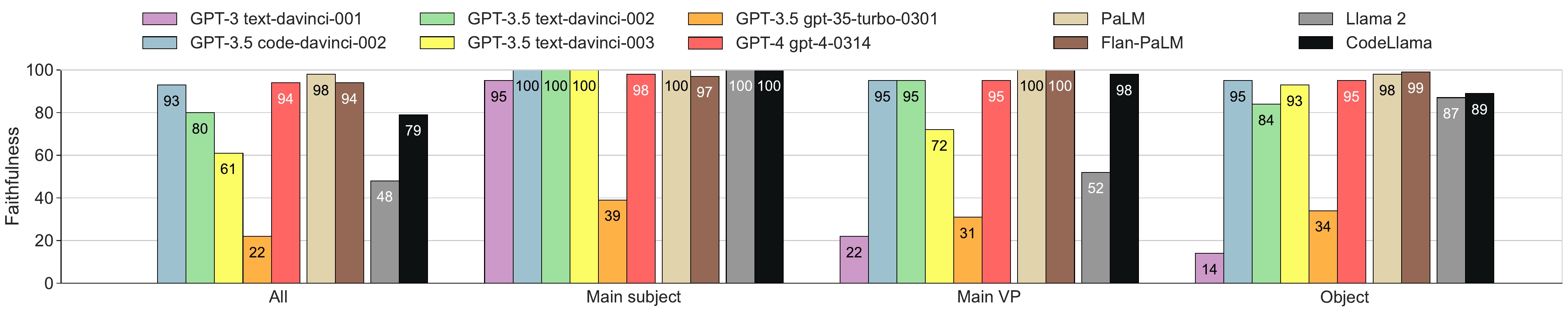}
    \vspace{-0.15cm}
    \caption{Reasoning accuracies and faithfulness scores for various models and syntactic components using the Code CoT prompt on question formation. ``All'' refers to the proportion of examples where models correctly label each syntactic component, or where outputs are entirely faithful to the model's generated reasoning.}
    \label{fig:qf_reasoningacc_faithfulness}
\end{figure*}

\begin{figure*}
    \centering
    \includegraphics[width=0.9\linewidth]{syntax_icl/figures/faithfulness/legend.png}
    \vspace{0.05cm}
        
    \includegraphics[trim={0 0 0 2.9cm},clip,width=0.9\linewidth]{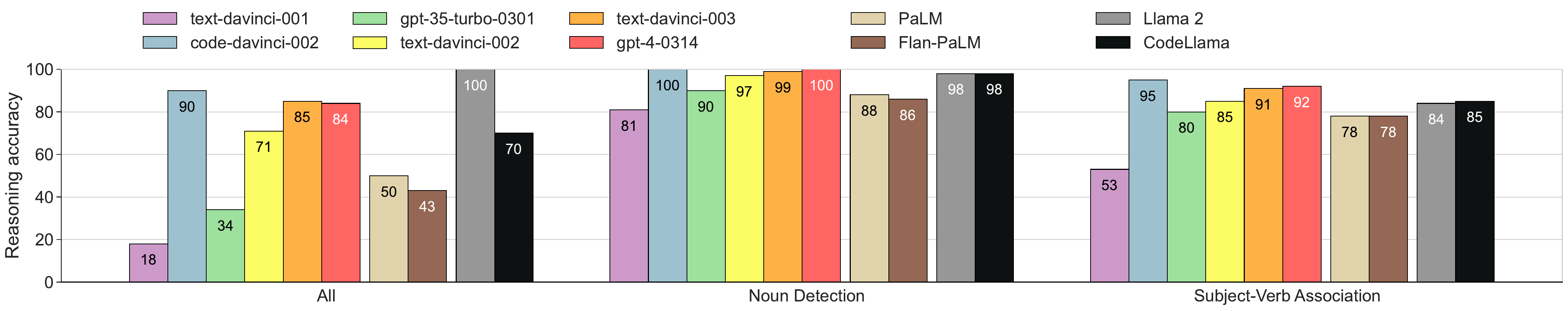}
    
    \includegraphics[trim={0 0 0 2.9cm},clip,width=0.9\linewidth]{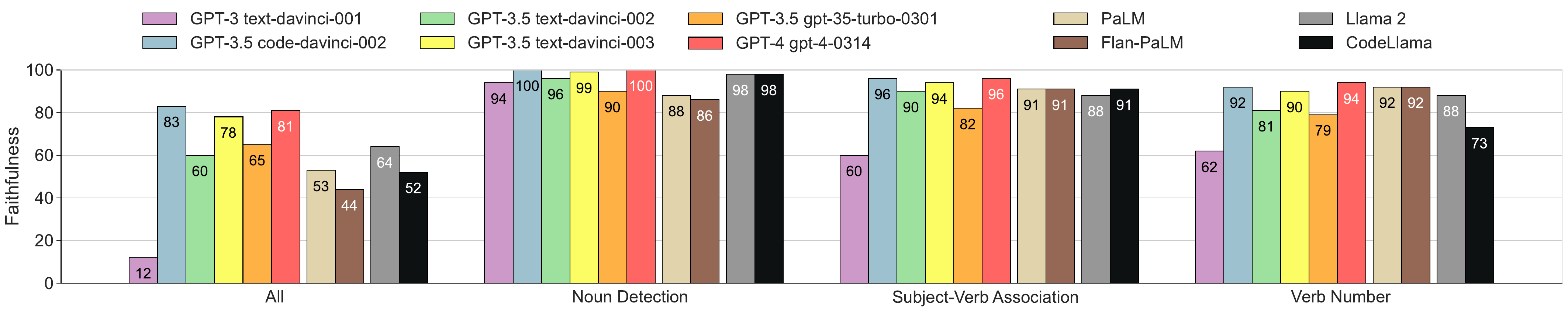}
    \vspace{-0.15cm}
    \caption{Reasoning accuracies and faithfulness scores for various models and syntactic components using the Code CoT prompt on tense reinflection. ``All'' refers to the proportion of examples where models correctly label each syntactic component, or where outputs are entirely faithful to the model's generated reasoning.}
    \label{fig:tense_reasoningacc_faithfulness}
\end{figure*}

Here, we present reasoning accuracies and faithfulness scores for all reasoning components, models, and tasks. See Figure~\ref{fig:qf_reasoningacc_faithfulness} for results on question formation, and Figure~\ref{fig:tense_reasoningacc_faithfulness} for results on tense reinflection.

\section{Syntactic Generalization in Text Classification: Prompts and Full Results}\label{app:hans}
The analyses reported in the main text found that LLMs leverage syntactic information to varying extents when taught a task via ICL, and that code pre-training improves OOD generalization. In this section, In this section, we test if these trends are specific to tasks where models must generate transformed versions of their inputs, or if they also extend to classification tasks. We leverage the Heuristic Analysis for NLI Systems (\textsc{Hans}; \citealp{mccoy-etal-2019-right}) dataset, which consists of Natural Language Inference (NLI) examples designed to diagnose reliance on syntactic heuristics. One typically trains the model on the training set of another NLI dataset, such as MNLI \citep{williams-etal-2018-broad} or RTE \citep{dagan-etal-2006-rte}, and then uses \textsc{Hans} as an evaluation set to measure whether models rely on syntactic heuristics. \citet{si2023prompting} evaluate RoBERTa and GPT-3 on \textsc{Hans} after fine-tuning on MNLI; they find that both models perform similarly, suggesting that scale may not be enough for models to overcome syntactic heuristics. We extend this analysis to a wider variety of LLMs and to chain-of-thought prompting, and further break down performance by label.

\begin{table*}[t]
    \centering
    \begin{subtable}{\linewidth}\centering
    \resizebox{0.8\linewidth}{!}{
    \begin{tabular}{lrrrrrr}
    \toprule
    & \multicolumn{2}{c}{Lexical Overlap} & \multicolumn{2}{c}{Subsequence} & \multicolumn{2}{c}{Constituent} \\\cmidrule(lr){2-3}\cmidrule(lr){4-5}\cmidrule(lr){6-7}
    \textbf{Model} & No CoT & Verbal CoT & No CoT & Verbal CoT & No CoT & Verbal CoT \\\midrule
    GPT-3 \texttt{text-davinci-001} & 59 & 100 & 55 & 97 & 59 & 99 \\
    GPT-3.5 \texttt{code-davinci-002} & 81 & 99 & 99 & 96 & 96 & 96 \\
    GPT-3.5 \texttt{text-davinci-002} & 84 & 98 & 67 & 99 & 64 & 95 \\
    GPT-3.5 \texttt{text-davinci-003} & 96 & 99 & 100 & 100 & 96 & 100 \\
    GPT-3.5 \texttt{gpt-3.5-turbo-0301} & 81 & 16 & 69 & 3 & 96 & 6 \\
    GPT-4 \texttt{gpt-4-0314} & 94 & 98 & 100 & 96 & 98 & 98 \\
    \midrule
    PaLM & 94 & 98 & 99 & 100 & 100 & 100 \\
    Flan-PaLM & 99 & 100 & 100 & 100 & 100 & 100 \\
    \midrule
    Llama~2 & 84 & 93 & 98 & 100 & 99 & 100 \\
    CodeLlama & 80 & 65 & 95 & 50 & 91 & 89 \\
    \bottomrule
    \end{tabular}}
    \caption{Entailment}
    \end{subtable}

    \vspace{0.5cm}
    
    \begin{subtable}{\linewidth}\centering
    \resizebox{0.8\linewidth}{!}{
    \begin{tabular}{lrrrrrr}
    \toprule
    & \multicolumn{2}{c}{Lexical Overlap} & \multicolumn{2}{c}{Subsequence} & \multicolumn{2}{c}{Constituent}\\\cmidrule(lr){2-3}\cmidrule(lr){4-5}\cmidrule(lr){6-7} 
    \textbf{Model} & No CoT & Verbal CoT & No CoT & Verbal CoT & No CoT & Verbal CoT \\\midrule
    GPT-3 \texttt{text-davinci-001} & 86 & 0 & 76 & 0 & 85 & 6 \\
    GPT-3.5 \texttt{code-davinci-002} & 99 & 94 & 70 & 48 & 74 & 35 \\
    GPT-3.5 \texttt{text-davinci-002} & 98 & 84 & 86 & 48 & 88 & 38 \\
    GPT-3.5 \texttt{text-davinci-003} & 99 & 99 & 50 & 34 & 61 & 11 \\
    GPT-3.5 \texttt{gpt-3.5-turbo-0301} & 82 & 5 & 27 & 11 & 9 & 12 \\
    GPT-4 \texttt{gpt-4-0314} & 94 & 92 & 48 & 46 & 49 & 35 \\
    \midrule
    PaLM & 97 & 94 & 57 & 57 & 27 & 13 \\
    Flan-PaLM & 78 & 96 & 36 & 41 & 3 & 2 \\
    \midrule
    Llama~2 & 99 & 91 & 78 & 51 & 27 & 11 \\
    CodeLlama & 96 & 100 & 56 & 89 & 55 & 47 \\
    \bottomrule
    \end{tabular}}
    \caption{Non-entailment}
    \end{subtable}
    \caption{Accuracies on \textsc{Hans}. Scores are split by label: test examples whose label is ``entailment'' are presented in the top table (a), while examples whose label is ``non-entailment'' are presented in the bottom table (b). High scores on entailment coupled with low scores on non-entailment signify that the model relies on the syntactic heuristic to predict the label.}
    \label{tab:accs_hans}
\end{table*}

\textsc{Hans} disagnoses reliance on three particular heuristics: lexical overlap, subsequence, and constituent heuristics. A model relying on \textbf{lexical overlap} heuristics assumes that two sentences with high word overlap entail each other. For example, although ``The actor paid the doctor'' means the opposite of ``The actor was paid by the doctor'', a model could assume these are entailed because of the high amount of word overlap between the two sentences. A model relying on \textbf{subsequence} heuristics assumes that a hypothesis is entailed if it is a subsequence in the premise. For example, although ``The actor near the doctor danced'' does not imply that ``The doctor danced'', a model may assume they are entailed because the latter sentence is fully contained in the first. Finally, a model relying on \textbf{constituent} heuristics assumes that a hypothesis is entailed if it is a subtree of the premise. For example, ``If the artist slept, the doctor ran'' does not imply that ``The artist slept'', but a model relying on this heuristic would assume that it is entailed. Models that systematically rely on any of these heuristics are expected to obtain 100\% accuracy on examples where the label is ``entailment'' because the heuristics happen to make the correct prediction; however, models that rely on these heuristics will also obtain 0\% accuracy on examples where the label is ``non-entailment''.

We evaluate our LLMs on HANS in the ICL setting. The 8 ICL exemplars are drawn from MNLI, and the test examples are from HANS. Our results are based on a uniform subsample of 100 examples per heuristic \emph{and} per label. We test across 3 syntactic heuristics, and there are 2 labels (entailment and non-entailment); thus, we have 600 test examples in total. We use No CoT prompts and Verbal CoT prompts similar to those we depict in Figure~\ref{fig:prompt_formats}. For example:

\vspace{-2ex}

\ex. \textbf{No CoT}
    \a. \emph{Input:}\newline Q: There are many regulations in place that will reduce air emissions from electric power generation. Is it definitely true that There are a lot of regulations in place that reduce emissions? \newline A: Yes \newline\newline \emph{...(up to 7 more exemplars from MNLI)} \newline\newline Q: The lawyer was advised by the actor. Is it definitely true that The lawyer advised the actor?\newline A:
    \b. \emph{Output:}\newline Yes

\vfill\eject

\ex. \textbf{Verbal CoT}
    \a. \emph{Input:}\newline
    Q: There are many regulations in place that will reduce air emissions from electric power generation. Is it definitely true that There are a lot of regulations in place that reduce emissions? \newline A: Regulations is the agent of both sentences. Emissions is the object inside the clauses in both sentences. Reduce is inside a clause in both sentences. The answer is Yes. \newline\newline \emph{...(up to 7 more exemplars from MNLI)} \newline\newline Q: The lawyer was advised by the actor. Is it definitely true that The lawyer advised the actor?\newline A:
    \b. \emph{Output:}\newline The lawyer is the subject of the first sentence. The actor is the subject of the second sentence. Advised is the verb in the first sentence. Was advised is the verb in the second sentence. The answer is Yes.

Our results (Table~\ref{tab:accs_hans}) suggest that LLMs are susceptible to syntactic heuristics, but to a lesser extent than smaller-scale LMs fine-tuned on MNLI \citep{mccoy-etal-2019-right}. Most models achieve close to 100\% accuracy on examples where the gold label is ``entailment'', as expected. However, on ``non-entailment'' examples, variance is very high: all models (except GPT-3.5 (Turbo)) achieve high accuracies on lexical overlap examples using both prompts, suggesting that they do \emph{not} succumb to the lexical overlap heuristic when generalizing. For other syntactic heuristics, however, non-entailment scores tend toward 0--50\%. This suggests that models rely on subsequence and constituent heuristics when guided via ICL, though the extent to which they rely on the heuristic depends on the prompt format. Unlike in the transformation tasks, here code pre-training does not seem to lend a significant advantage: there is no consistent significant gain (nor loss) on entailment or non-entailment examples in models pre-trained on code.

We also find that \textbf{chain-of-thought reasoning makes models more prone to relying on syntactic heuristics.} With Verbal CoT, scores on entailment examples tend to increase across models and across syntactic heuristic types. On non-entailment examples, however, it reduces scores to near-random-chance or near-zero. This pattern suggests that Verbal CoT pushes models to prefer the syntactic heuristics more strongly.

\end{document}